\documentclass{article} 
\usepackage[final]{colm2025_conference}

\usepackage{microtype}
\usepackage{hyperref}
\usepackage{url}
\usepackage{booktabs}

\usepackage{lineno}

\definecolor{darkblue}{rgb}{0, 0, 0.5}
\hypersetup{colorlinks=true, citecolor=darkblue, linkcolor=darkblue, urlcolor=darkblue}

\usepackage{microtype}

\usepackage{inconsolata}

\usepackage{graphicx}

\usepackage[utf8]{inputenc} 
\usepackage[T1]{fontenc}    
\usepackage{url}            
\usepackage{booktabs}       
\usepackage{amsfonts}       
\usepackage{nicefrac}       
\usepackage{microtype}      
\usepackage{xcolor}         
\usepackage{multirow}
\usepackage{graphicx}
\usepackage{pifont}
\usepackage{makecell}

\newcommand{\fref}[1]{Figure~\ref{#1}}
\newcommand{\tref}[1]{Table~\ref{#1}}
\newcommand{\sref}[1]{\S\ref{#1}}

\usepackage{microtype}
\usepackage{graphicx}
\usepackage{amsmath}
\usepackage{bbm}
\usepackage{tcolorbox}

\usepackage{enumitem}
\usepackage{CJKutf8}

\usepackage[]{todonotes}

\newcounter{obssection}

\title{SitEmb-v1.5: Improved Context-Aware Dense Retrieval\\ for Semantic Association and Long Story Comprehension}

\author{Junjie Wu$^1$, Jiangnan Li$^2$,
Yuqing Li$^3$, Lemao Liu$^2$, Liyan Xu$^2$, Jiwei Li$^4$\\
\bfseries Dit-Yan Yeung$^1$, Jie Zhou$^2$, Mo Yu$^2$\hspace{-0.1em}\thanks{Corresponding Author.} \\
$^1$HKUST\quad $^2$WeChat AI, Tencent \quad $^3$IIE-CAS\quad $^4$Zhejiang University \\
\texttt{junjie.wu@connect.ust.hk\quad\{jiangnanli,moyumyu\}@tencent.com} \\
}

\begin{document}

\ifcolmsubmission
\linenumbers
\fi

\maketitle

\begin{abstract}
Retrieval-augmented generation (RAG) over long documents typically involves splitting the text into smaller chunks, which serve as the basic units for retrieval.
However, due to dependencies across the original document, contextual information is often essential for accurately interpreting each chunk.
To address this, prior work has explored encoding longer context windows to produce embeddings for longer chunks.
Despite these efforts, gains in retrieval and downstream tasks remain limited.
This is because (1) longer chunks strain the capacity of embedding models due to the increased amount of information they must encode, and (2) many real-world applications still require returning localized evidence due to constraints on model or human bandwidth.

We propose an alternative approach to this challenge by representing short chunks in a way that is conditioned on a broader context window to enhance retrieval performance -- \emph{i.e.}, \textbf{situating} a chunk's meaning within its context. We further show that existing embedding models are not well-equipped to encode such situated context effectively, and thus
introduce a new training paradigm and develop the \emph{situated embedding models (\textbf{SitEmb})}. 
To evaluate our method, we curate a book-plot retrieval dataset specifically designed to assess situated retrieval capabilities. On this benchmark, our SitEmb-v1 model based on BGE-M3 substantially outperforms state-of-the-art embedding models, including several with up to 7-8B parameters, with only 1B parameters.
Our 8B SitEmb-v1.5 model further improves performance by over 10\% and shows strong results across different languages and several downstream applications.\footnote{\url{https://huggingface.co/SituatedEmbedding}}

\end{abstract}

\section{Introduction}
\label{sec:introduction}
Text embedding models~\citep{wang2024improving,sturua2024jinaembeddingsv3multilingualembeddingstask,nussbaum2024nomic,hu2025kalm} encode textual inputs into vector spaces. These models enable efficient semantic representation and matching, thus are foundational to many applications involving retrieval-augmented generation (RAG)~\citep{lewis2020retrieval}, such as code generation~\citep{wang2024coderag,miao2024integrating}, reference generation~\citep{wu-etal-2025-ref}, and personal AI assistants~\citep{notebookLM}.

In these tasks, candidate documents are typically segmented into smaller chunks to facilitate efficient processing. 
However, since documents often exhibit a narrative or logical flow, the meaning of each chunk is highly dependent on its surrounding context. 
This highlights the need for text embeddings that capture broader contextual information to enable \textbf{context-aware retrieval}.

One straightforward approach to this issue is to increase chunk size, allowing each chunk to capture more information.
This has motivated a wave of recent work on supporting long input sequences in embedding models, either by designing efficient bidirectional models~\citep{bge-m3,sturua2024jinaembeddingsv3multilingualembeddingstask,nussbaum2024nomic}, or by repurposing powerful unidirectional pre-trained LLMs as embedding generators~\citep{li2023towards,wang2024improving,moreira2024nv,LinqAIResearch2024}. These models can produce embeddings for sequences of up to 8,192 tokens or more.

\begin{figure}
    \centering
    \includegraphics[width=0.7\textwidth]{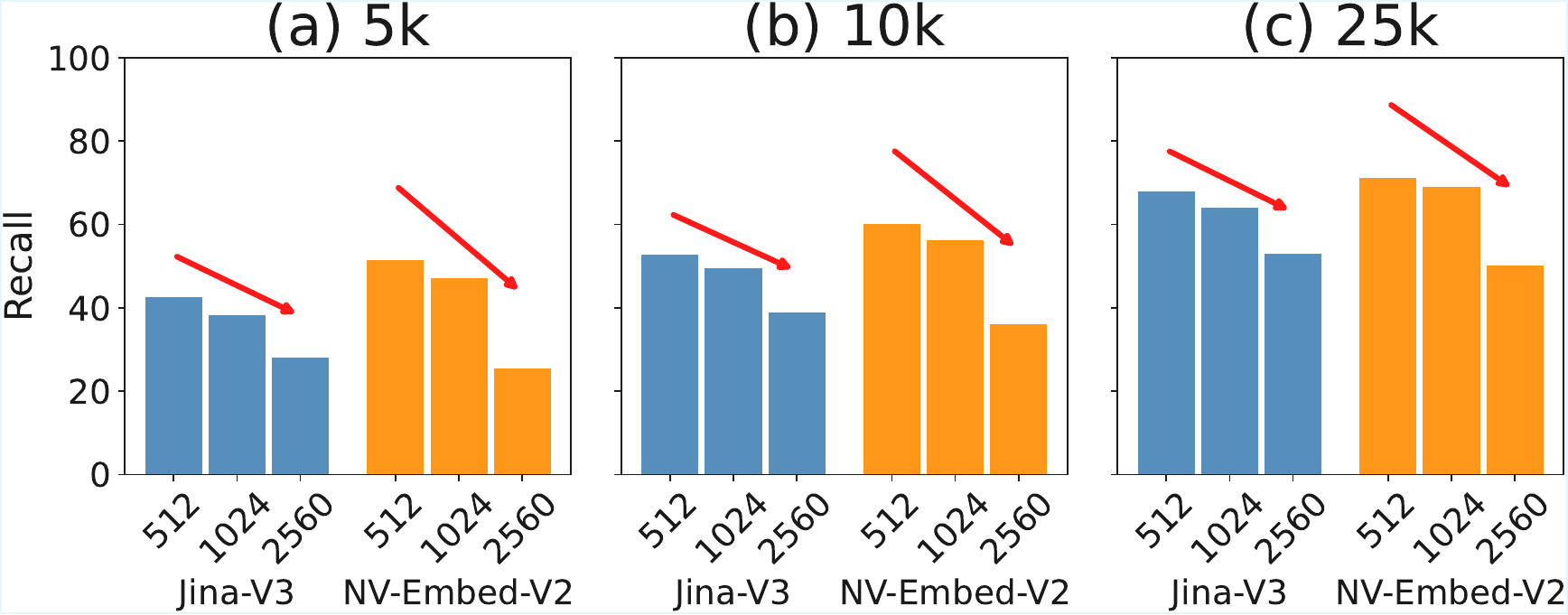}
    \caption{Comparison of the same embedding models that return the same lengths of texts with different chunk sizes on our evaluation task (\sref{sec:dataset}). X-axis refers to chunk sizes. For example, when the return text is 5k and the chunk size is 1,024, the retriever returns top-5 chunks.
    } 
    \vspace{-0.2in}
    \label{fig:teaser}
\end{figure}

However, it is often observed that simply \emph{enabling longer input windows does not necessarily lead to better embeddings}.
A key reason lies in the \textbf{limited capacity} of embedding vectors -- embedding models must compress the information in the input text into a single vector. Intuitively, the longer the input chunk, the more information it contains, and the more long-range dependencies across arbitrary pairs of chunks within a document it needs to capture. This increases the likelihood of critical information loss during compression.
Existing models are trained by merely extending the context window, without explicitly learning how to represent such distributed contextual relationships, which leads to a counterintuitive outcome: applications built on long-chunk embeddings often underperform those using short-chunk embeddings, despite the latter discarding more contextual information.
Figure~\ref{fig:teaser} illustrates this effect on a book plot retrieval task~\citep{xu2024fine}.
When the same total length of text (5k/10k/25k tokens) is retrieved using the same embedding model (Jina or NV-Embed), recall consistently decreases as the documents are segmented into longer chunks.

Given the aforementioned challenges, we propose an alternative approach to context-aware embedding: directly incorporating the broader context surrounding each short chunk into its chunk embedding. This allows the model to account for how the chunk is situated within the original document, enabling more contextually informed embeddings.
In other words, we aim to situate a chunk's meaning within its broader context \citep{yu2023personality,xu2024fine} during the embedding process.
We refer to this approach as \textbf{situated embedding}.
By doing so, we alleviate the issue of capacity limitations: during encoding, the model only needs to identify and integrate context that is relevant to the target chunk, which is a more tractable task than modeling all dependencies across an extended input window.

Building on the idea of situated embedding, we first investigate whether existing embedding models can effectively generate situated embeddings. However, we find that \emph{situated embedding cannot be achieved simply by prompting existing embedding models}, as demonstrated in \sref{sec:study1}. To address this limitation, we develop a dedicated situated embedding model specifically designed to handle this scenario.
We achieve this through two techniques:
(1) \emph{Constructing context-dependent training instances} using publicly available user-annotated book notes. Platforms such as Douban\footnote{\url{https://book.douban.com/}} allow users to write notes anchored to particular book segments.
We treat the note as a query and the anchor text as groundtruth, framing a retrieval task with \textasciitilde 1.6M query-candidate pairs.
As user notes typically reflect the contextual understanding of surrounding context, it makes context-aware embeddings beneficial for this retrieval task.
(2) \emph{Promoting context usage through residual learning}.
In many cases, a chunk alone may offer partial (usually ambiguous) clues about its relevance to the query, allowing models to exploit shortcuts.
To counter this, we employ a residual architecture where the situated embedding model is trained to resolve the residual from a baseline chunk-only embedding model. This encourages the model to focus on the additional contextual information.

To evaluate models' context-aware embedding capability, we curate a book-plot retrieval task following~\citep{xu2024fine}, which has been verified by previous work for its guaranteed requirement of context-aware embedding capability.
Experiments demonstrate our situated embedding model's superior performance over all the state-of-the-art embedding models, including those with up to 7B parameters and with massive pre-training. 
We further illustrate the generalizability of the trained models on the \emph{Recap Snippet Identification} task~\citep{li2024previously}, a distinct task in book understanding beyond standard query-candidate retrieval, and several downstreaming story understanding tasks such as QA and claim verification that require RAG.

\section{Our Situated Embedding Model}
\label{sec:method}

We develop the first model that generates high-quality situated embeddings as follows:

\paragraph{Training Data Construction}

We construct two sets of training data in English and Chinese, corresponding to different usage scenarios of retrieval, long-story comprehension, and semantic association. For the story comprehension purpose, we follow \citet{mou2021narrative} and build data based on NarrativeQA~\citep{kovcisky2018narrativeqa}. For association-oriented training, we draw on book notes following prior work~\citep{yu2023personality,zhou2025essence}. Book notes are particularly suitable for this purpose, as they capture human readers’ divergent thinking while engaging with a paragraph, thereby revealing associations between notes and the corresponding text. The construction of this book-note training data is detailed as follows.

We collect the notes and their associated anchor texts for \textasciitilde 100 most popular books according to Douban.
We treat each user note as a query and its corresponding anchored segment as a chunk, resulting in 1,614,007 query–chunk pairs. 
We reserve all the query–chunk pairs from our evaluation books and randomly select 1000 pairs for early-stopping.

Given a query–chunk pair, we define the situated context of the chunk as \emph{a sequence of its surrounding sentences, including the chunk itself}. Specifically, we use the user-underlined texts anchored to their query as the chunk's situated context, as these texts naturally align with our definition. Due to variations in user behavior, the lengths of these situated contexts range from 37 tokens to several thousand, making our trained model robust to a wide range of context lengths.

After this process, each chunk will be contained within one such segment, and we regard the segment as the situated context of the chunk.

\paragraph{Residual Learning to Promote Situated Context Usage}
Prior studies, such as~\citep{ettinger2020bert}, have shown that BERT-based models often rely on shallow heuristics or partial, ambiguous clues when matching texts. This behavior hinders the model’s ability to fully comprehend the entire input, which potentially explains why existing embedding models struggle to utilize long contextual information.
To address this limitation, we adopt a residual learning framework~\citep{he2016deep}, in which a situated embedding model is trained to resolve the residual from a baseline chunk-only embedding model, thereby equipping the trained model with a deeper understanding of situated context.

Specifically, we maintain two models, a baseline model $\Theta^b$ that embeds the chunk only and a situated model $\Theta^s$ that embeds the chunk situated within the context.
For each query-chunk pair in the training data, we treat the chunk as the positive sample, and randomly sample 10 other chunks from the remaining chapters of the same book as negative samples. 
A query is embedded as $\mathbf{\tilde{q}}=\mathbf{q}^b+\mathbf{q}^s$, where $\mathbf{q}^b$ and $\mathbf{q}^s$ are embedding vectors from $\Theta^b$ and $\Theta^s$, respectively.
Similarly, a chunk is embedded as $\mathbf{\tilde{c}}=\mathbf{c}^b+\mathbf{c}^s$.
The training loss on each query-chunk pair can then be defined as:
\begin{equation}
\small
\begin{aligned}
\mathcal{L}(\Theta^{b}, \Theta^{s}) = \frac{1}{N} \sum_{i=1}^{N=10} \max\big(&0,\ \gamma + \text{sim}(\mathbf{\tilde{q}}_j, \mathbf{\tilde{c}}^-_{j,i}) \\
&- \text{sim}(\mathbf{\tilde{q}}_j, \mathbf{\tilde{c}}^+_j)\big),
\end{aligned}
\label{eq1}
\end{equation}
where $i$ is the index of negative chunk. See Appendix~\ref{app:training} for details of the training process.

\section{Evaluation Dataset}
\label{sec:dataset}

\citet{xu2024fine} repurpose instances from the PlotRetrieval dataset~\cite{xu2024plot} to support the task of contextual retrieval. Their work focuses on a single book, demonstrating that incorporating a graph-based representation of the book can improve local chunk retrieval. This finding highlights that the plot retrieval task inherently requires situated understanding and retrieval capabilities.

Following their work, we repurpose the PlotRetrieval dataset into a chunk-level retrieval task and expand the number of evaluation books from 1 to 7. Specifically, we filter books that are too short (i.e., $\le$100{,}000 tokens), as they can typically be processed in a single input window and therefore diminish the utility of RAG. We also exclude books with too few user notes, as well as less popular versions on the reading platform, which tend to have 
less diverse note styles.

This filtering process results in 7 evaluation books containing 1,394 diverse queries, which together constitute the \emph{Book Plot Retrieval} task (see Appendix~\ref{app:books} for details on each selected book).
When constructing the situated context for each chunk in the \emph{Book Plot Retrieval} task, we first partition the chunk’s corresponding book into segments of consecutive sentences util the length reaches 200 tokens. We then sequentially group every 16 consecutive segments. This grouped context serves as the situated context of each chunk inside the group. During evaluation, we report Recall@10, Recall@20, and Recall@50 as the primary metrics. 

\paragraph{Remark} 
Our plot retrieval task captures an important real-world scenario where localized retrieval results are essential. When users of an online reading app try to recall a plot, they typically lack the time or patience to read through long passages. In our setting, each retrieved segment corresponds to about 2–3 pages as displayed in a mobile or digital reading app, aligning with this user requirement.
However, while humans can mentally connect these short segments to the broader narrative, the content alone is often insufficient due to missing context, highlighting the need for situated embedding techniques.

\begin{table*}[t]
  \centering
  \small
   \setlength{\tabcolsep}{1.4mm}{
  \begin{tabular}{lc|ccc|ccc|ccc}
    \toprule
    \multirow{3}*{\textbf{Model}} & \multirow{3}*{\textbf{Size}}  & \multicolumn{3}{c}{\textbf{Chunk-Only}} & \multicolumn{3}{c}{\textbf{+ Situated Context}} & \multicolumn{3}{c}{\textbf{+ Situated Summ.}} \\
    \cmidrule(lr){3-5} \cmidrule(lr){6-8} \cmidrule(lr){9-11}
  && @10 & @20 & @50 & @10 & @20 & @50 & @10 & @20 & @50\\
  \midrule[0.5pt]
  M3 & 0.5B & 42.55 & 53.33 & 66.51 & 9.48 & 15.68 & 24.70 &41.83 &53.94 &67.78 \\
  Jina-v3 & 0.5B & 42.65 & 52.67 & 67.92 & 34.10 & 44.27 & 58.33 & 45.48 & 56.12 & 69.90 \\
  \midrule[0.5pt]
  E5-Mistral & 7B & 43.18 & 51.93 & 66.65 & 14.77 & 24.49 & 32.68 & 44.58 & 54.32 & 68.27 \\
  GTE-Qwen2 & 7B & 46.19 & 55.79 & 71.15 & 19.01 & 29.77 & 49.34 & 42.44 & 49.88 & 60.85 \\
  NV-Embed-v2 & 7B &  51.38 & 60.11 & 71.16 & 21.01 & 30.20 & 42.18 & 49.25 & 58.33 & 70.85 \\
  Qwen3-Embedding & 8B &  51.58 & 61.32 & 73.47 & 48.01 & 58.91 & 71.86 &  51.20 & 61.63 & 76.10  \\
  \midrule[0.5pt]
  voyage-context-3 & unk & 58.54 & 68.47 & 79.09 & 60.46 & 69.50 & 82.19  & 61.37 & 70.36 & 82.84\\
  \midrule[0.5pt]
  SitEmb-v1-M3 (ours) & 1B & 48.79 & 58.45 & 73.29 & 50.85 & 60.57 & 76.40 &  51.47 & 63.69 & 76.76 \\
  SitEmb-v1.5-Qwen3 (ours) & 8B & 61.66 & 69.06 & 79.54 &  63.32	& 72.75 & 83.59 & 61.26 &	70.02 & 81.57\\
  \quad + book note data & 8B & \bf 66.81 &	\bf 74.36 & \bf 84.32 &\bf  68.98	&\bf 79.32 &\bf 86.68 & \bf 67.81 &	\bf 76.37 & \bf 84.37\\
    \bottomrule
  \end{tabular}
  \caption{{Recall results on NDP-v1. The maximum length is set to 8,192. Best results of each setting are boldfaced.}}
   \vspace{-0.2in}
  \label{tab:ndp_exp}
   }
\end{table*}

\section{Study I: Analysis of Existing Models on Generating Situated Embeddings}
\label{sec:study1}
As the first step, we investigate the necessity of training a situated embedding model. That is, \emph{are existing long-context embedding models capable of generating good situated embeddings?}

\paragraph{Setup} We investigate this question on the NDP-v1 book in our evaluation dataset.
Following our approach described in~\sref{sec:method}, we use the 16 surrounding chunks of each chunk to construct its situated context.
We compare the following models\footnote{The models are selected based on their strong performance on the MTEB benchmark~\citep{muennighoff2023mteb}. }: 1) \emph{Long-context BERT models,} including BGE-M3~\citep{bge-m3} and Jina-v3 (\texttt{Jina-Embeddings-v3})~\citep{sturua2024jinaembeddingsv3multilingualembeddingstask}. 2) \emph{LLM-based embedding models}: E5-Mistral (\texttt{E5-Mistral-7b-Instruct})~\citep{wang2024improving}, GTE-Qwen2 (\texttt{GTE-Qwen2-7b-Instruct})~\citep{li2023towards}, NV-Embed-v2~\citep{lee2024nv} and Qwen3-Embedding-8B~\citep{zhang2025qwen3}. 3) \emph{Our trained situated model} from~\sref{sec:method}, including the v1 model based on M3 and the v1.5 model based on Qwen3.
For reference, we also compare with the concurrent work on the most advanced commercial late-chunking model \texttt{voyage-context-3} \citep{voyage2025context3}.
Check Appendix~\ref{app:run_model} for additional details on model usages.

\paragraph{Results} 
Table~\ref{tab:ndp_exp} presents the evaluation results, from which we draw the following conclusions:

\noindent$\bullet$\emph{ Existing models does not have zero-shot situated embedding capability.}
When enhancing the contexts to chunks, the performance of all the existing models degrades significantly (\emph{i.e.}, comparing columns of \emph{+Situated Context} and \emph{Chunk-Only}).
Note that the length of the situated context is well within their claimed maximum context window sizes.
In contrast, our situated embedding model can effectively leverage contextual information, and largely surpasses the much larger 7B baselines.

\noindent$\bullet$\emph{ The poor results partly sourced from limitation in understanding long inputs.} The failure of producing situated embeddings is partly from the existing models' (actual) insufficiency of handling long inputs.
To see this, we in addition compare with the \textbf{LLM-generated situated summaries} approach~\citep{anthropic1}, which prompts an LLM to generate a concise summary that reflects how a chunk is situated within its broader context as the contextual information. 
We ask GPT-4o~\citep{gpt4o} to generate the situated summaries and use them in the same way like the situated contexts.
Note that we use this setting only for reference, because it does not make a fair comparison due to the involvement of a much stronger model in the pipeline with high computational cost.

From the results, all the models suffer from a much smaller degrade when using the summaries instead of original situated contexts, while M3, Jina, E5 and Qwen3 have their results slightly increased.
This reflects the fact that the baselines fail to situate the target chunk within long contexts.
In comparison, our approach can achieve performance boost for both types of contextual inputs.

\section{Study II: Analyzing the Robustness of Our SitEmb Models}

In this study, we examine two aspects crucial to real-world applications: (1) whether our SitEmb models learn to generalize to new books rather than rely on memorization, and (2) whether they are robust to variations in situated context length.

\paragraph{The Impact of Training-Test Book Overlap}
A key concern in evaluating pre-trained language models, including embedding models, is whether models benefit unfairly from training-test overlap.
Demonstrating that such overlap does not drive results is particularly important, because model training cannot anticipate all future downstream uses; it is therefore impractical to proactively filter training data against every possible evaluation or user scenario. 

To verify the validity of our evaluation, we constructed a controlled experiment that modifies the training data from NarrativeQA with and without any version of the NDP books,\footnote{We use NarrativeQA rather than book-note data, since the queries in the plot retrieval task are originally derived from book notes so the test books have to be removed, making such an overlap experiment implausible in that setting.} and then evaluated their performance on NDP-v1. As shown in Table~\ref{tab:analysis_book_overlap}, models exposed to the test books during training exhibit no measurable performance gain, indicating that training-test overlap does not materially affect our results.

\begin{table}[t]
  \centering
  \small
  \begin{tabular}{lccc}
    \toprule
     \multirow{2}{*}{\textbf{Setting}} &\multicolumn{3}{c}{\textbf{Recall}}\\
    \cmidrule(lr){2-4}
    & @10 & @20 & @50 \\
    \midrule[0.5pt]
     w/ NDP & 68.98 &	79.32	& 86.68 \\ 
     w/o NDP &69.60&78.72&87.04 \\
    \bottomrule
  \end{tabular}
  \caption{{Study the impact of training-test book overlap on the NDP-v1 task. We experiment with the Sit-Qwen3 with book-note setting.}}
  \label{tab:analysis_book_overlap}
\end{table}

\paragraph{Robustness to Context Length}
We evaluate the sensitivity of our trained model to variations in context length.  Experiment is conducted on the NDP-v1 book used in~\tref{tab:ndp_exp}, varying the number of sentences per chunk and measuring recall scores with our SitEmb-v1-M3 and the SitEmb-v1.5-Qwen models, both trained with book note data. The results in~\tref{tab:build situated} demonstrate that the model maintains stable performance across different context lengths. Our choice of 16-segment groups strikes a balance between efficiency and accuracy.

\begin{table}[t]
  \centering
  \small
  \setlength{\tabcolsep}{2mm}
  \begin{tabular}{c|ccc|ccc}
    \toprule
    \multirow{3}{*}{\textbf{Situated Context Length}} & \multicolumn{3}{c}{\textbf{SitEmb-v1 Recall}} & \multicolumn{3}{c}{\textbf{SitEmb-v1.5 Recall}} \\
    \cmidrule(lr){2-4}\cmidrule(lr){5-7}
    & @10 & @20 & @50 & @10 & @20 & @50\\
    \midrule[0.5pt]
    $[512,\ 800]$     & 51.63 & 61.22 & 74.48 & \underline{69.31} & \underline{77.51} & 86.09\\
    $[1024,\ 1600]$    & \bf 52.29 & \underline{61.23} & 74.15 & 68.76 & 76.24 & 85.78\\
    $[2048,\ 3200]$    & \underline{51.73} & \bf 61.56 & \underline{75.00} &{68.98}	&\bf 79.32 &\underline{86.68}\\
    $[4096,\ 6400]$   & 50.62 & 59.11 & \bf 75.16 &\bf 69.75 & 77.45 & \bf 86.83 \\
    $[8192,\ 12800]$   & 50.52 & 59.11 & 74.51 & 68.88 & 77.00 & 86.47\\
    \bottomrule
  \end{tabular}
  \caption{
    Recall results of our situated embedding models on NDP-v1 with various lengths of situated contexts. The listed lengths correspond to multiples (4/8/16/32/64) of the average segment range observed in books from the book plot retrieval task. The best results are \textbf{boldfaced} and the second best results are \underline{underlined}.
  }
  \label{tab:build situated}
\end{table}

\section{Study III: Contextual Retrieval on the Full Book Plot Retrieval Task}

We evaluate our situated embedding model on the full plot retrieval task to assess its effectiveness in enhancing contextual retrieval. 
We compared our models trained with the QA data (denoted as \emph{QA}) and book note data (semantic association, denoted as \emph{SA}).
The M3 model fails to improve with the QA training data thus the corresponding results are omitted.

As shown in Table~\ref{tab:main}, incorporating contextual information through situated embeddings significantly improves performance. Our SitEmb-v1-M3 model consistently outperforms chunk-only baselines without our training techniques.
The SitEmb-v1.5 models further boost performance by over 10\% when trained on QA data and over 15\% when trained on QA+SA data. Notably, both variants surpass the performance of the recent commercial late-chunking model voyage-context-3, and show clear advantages over their chunk-only variations.

To ensure that the gains of SitEmb-v1-M3 are not merely due to increased model capacity, we also train the same residual architecture on two chunk-only M3 models (Res-M3). It fails to yields improvements over the trained M3 (SA) baseline, indicating that the advantage of our method primarily come from the effective use of contextual information. In addition, training without the residual architecture (- Residual) leads to degraded performance compared to our full SitEmb-v1-M3, further supporting our training design.

\begin{table}[t]
  \centering
  \small
  \setlength{\tabcolsep}{1mm}{
  \begin{tabular}{llc|ccc}
    \toprule
    \multirow{2}{*}{\textbf{Setting}}  & \multirow{2}{*}{\textbf{Model}} & \multirow{2}{*}{\textbf{Size}} &\multicolumn{3}{c}{\textbf{Recall}}\\
    \cmidrule(lr){4-6}
    &&& @10 & @20 & @50 \\
    \midrule[0.5pt]
    \multirow{6}{*}{Chunk-only}& M3 (out-of-box) & 0.5B & 32.92& 41.46&55.85 \\ 
    & M3 (SA) & 0.5B &42.87 &52.91 &66.01 \\ 
    & \quad + Residual & 1B& 43.43&51.74 &65.51 \\
    & Qwen3 (out-of-box)& 8B& 43.57 & 52.94 &	69.48 \\
    & Qwen3 (QA)& 8B& 51.02 & 60.78 &	73.89 \\
    & Qwen3 (QA+SA) & 8B & 60.36 & 68.87 &	80.45 \\ 
    \midrule[0.5pt]
    Late Chunking & voyage-context-3 & unk & 51.39 & 60.89 & 73.70 \\ 
    \midrule[0.5pt]
    \multirow{4}{*}{Situated}& SitEmb-v1-M3 (SA) & 1B&{45.15} &{55.66} &{69.25} \\ 
    &\quad - Residual & 0.5B &43.85&54.98&68.93 \\
    & SitEmb-v1.5-Qwen3 (QA) & 8B& 53.57 & 63.56 & 78.64 \\ 
    & SitEmb-v1.5-Qwen3 (QA+SA) & 8B & \bf 63.03 & \bf 72.83 &	\bf 82.70 \\ 
    \bottomrule
  \end{tabular}
  }
  \caption{{Overall results on book plot retrieval. Check~\tref{tab:main_full_m3}} and 
 \ref{tab:main_full_qwen} for full results.}
  \vspace{-0.1in}
  \label{tab:main}
\end{table}

\begin{table}[t]
  \centering
  \small
  \setlength{\tabcolsep}{2.7mm}{
  \begin{tabular}{l|ccc}
    \toprule
    & \multicolumn{3}{c}{\bf Recap}\\
    \cmidrule(lr){2-4}
     \textbf{Model} & R@5 & P@5 & F1@5 \\
    \midrule
    Qwen3 & 33.0 & 46.6 & 37.9\\
    \midrule
       Qwen3 (QA) & 32.0 &	45.8 &	37.0\\
       Qwen3 (QA+SA) & 32.6 &	46.4 &	37.6\\
       \midrule
       SitEmb-v1.5-Qwen3 (QA)& 32.3&	46.6 & 37.4\\
       SitEmb-v1.5-Qwen3 (QA+SA) & \bf 33.6 & \bf 48.2 & \bf 38.9 \\
    \bottomrule
  \end{tabular}
  \caption{{Results on the recap task.}}
  \label{tab:recap_overall}
  }
\end{table}

\section{Study IV: Downstream Semantic Association Task  -- Recap Identification}

Next, we assess the generalizability of our situated embedding model on downstream applications that are not explicitly designed for contextual retrieval and contain only a limited portion of context-dependent examples.
In this section, we evaluate on a distinct task, \emph{Recap Snippet Identification task}~\citep{li2024previously}, which aims to identify recap passages for a given paragraph.
We following their setting of using top-5 retrieved passages.

\paragraph{Results}

Table~\ref{tab:recap_overall} presents the results.
Because this task differs substantially from our training data, it poses a challenging transfer setting. Consequently, our trained models without context usage show a slight performance drop compared to the original Qwen3 model.
However, since recap identification requires embedding capabilities beyond simple similarity matching, our models trained with SA data achieve better generalization to this task. In particular, the SitEmb-v1.5 (QA+SA) model outperforms all others by leveraging contextual information.
These results highlight the importance of enhancing semantic association capabilities in embeddings. 

\section{Study V: Downstream Long Story Comprehension Applications}
Finally, we evaluate on a variaty of story understanding tasks that requires processing inputs exceeding the length limits of many LLMs, including \emph{NarrativeQA}~\citep{kovcisky2018narrativeqa}, the multichoice QA task from \emph{$\infty$Bench}~\citep{DBLP:conf/acl/ZhangCHXCH0TW0024}, the newly release \emph{DetectiveQA}~\citep{xu2025detectiveqa}, the public subset of \emph{NoCha}~\citep{DBLP:conf/emnlp/KarpinskaTLGI24} and the LongStoryQA-large task from \emph{CLongEval}~\citep{qiu2024clongeval}.
These tasks cover different genres, both English and Chinese languages, and task types of free-form QA, multi-choice QA and claim verification.
We retrieve top-3/5/10 with the compared embedding models and use Qwen2.5-72B (4-bit quantized) model to generate the results.

\begin{table*}[t]
  \centering
  \small
  \resizebox{1.\columnwidth}{!}{
  \begin{tabular}{l|c|c|c|c|c}
    \toprule
    \multirow{3}{*}{\textbf{Model}}& {\bf NarrativeQA}& {\bf $\mathbf{\infty}$Bench-En.MC} & {\bf DetectiveQA} & \bf NoCha (Public)& \bf LongStoryQA-Large\\
    \cmidrule(lr){2-2}\cmidrule(lr){3-3}\cmidrule(lr){4-4}\cmidrule(lr){5-5}\cmidrule(lr){6-6}
      & F1 & Acc & Acc & Pair Acc & F1\\
       \midrule
       Qwen3 (out-of-box) &  27.5/30.8/32.2 & 75.1/80.4/86.0 & 62.5/68.7/73.2 & 42.9/41.3/46.0 & 52.7/57.9/61.2\\ 
       \midrule
       Qwen3 (QA)  & 29.5/31.9/32.4 & \textbf{83.0}/84.7/88.7 & 70.5/78.2/81.8 & 54.0/52.4/36.5& \textbf{58.3}/59.2/61.4\\ 
       SitEmb-v1.5-Qwen3 (QA)&\textbf{31.1}/\textbf{32.0}/\textbf{34.4} & \textbf{83.0}/\textbf{86.9}/\textbf{90.0} & \textbf{73.2}/\textbf{78.7}/\textbf{82.3} & 54.0/\textbf{55.6}/46.0 & 57.7/58.7/\bf 61.9\\ 
       SitEmb-v1.5-Qwen3 (QA+SA) & 29.4/31.6/31.8 & \textbf{83.0}/85.6/88.2 & 66.5/74.2/78.3 & \textbf{55.6}/52.4/\textbf{49.2} & 57.7/\textbf{59.4}/61.5\\
    \bottomrule
  \end{tabular}
  }
  \caption{{Results on the story QA tasks. We report results with top-3/5/10 retrieved chunks.}}
  \label{tab:qa_overall}
\end{table*}

\paragraph{Overall Results}
Table~\ref{tab:qa_overall} shows that our SitEmb-v1.5 model trained on QA data consistently outperforms its counterpart without situated embeddings, except on LongStoryQA. It also substantially outperforms the original Qwen3 embedding model, particularly on top-3 and top-5 results.
In comparison, our SitEmb model trained on QA+SA yields mixed results relative to the no-context model, but still shows advantages over the original Qwen3. This suggests that \emph{existing story comprehension tasks demand limited semantic association capability}, making the SitEmb (QA) model a well-balanced choice across diverse benchmarks.

One notable observation regarding performance degradation with larger retrieved context on NoCha is that, once the key plot is retrieved, additional context tends to consist mainly of distractors, causing an LLM with weaker reasoning ability to lose focus.
To verify this, we evaluated the advanced Gemini-2.5-Flash under our QA+SA setting, achieving top-3/5/10 pair accuracies of 55.6/57.1/57.1 without degradation. This confirms that the necessary evidence is saturated within the top-5 results.

\paragraph{Fine-Grained Evaluation on Retrieval Results}
Finally, we perform a fine-grained evaluation on the DetectiveQA dataset, which includes human-annotated evidence locations. Specifically, DetectiveQA provides two types of evidence annotations: \emph{Answer Evidence}, which refers to the text span that directly yields the answer; and \emph{Clue Evidence}, which refers to the supporting information that connects the evidence to the correct answer, mirroring the logical reasoning steps a detective would follow to solve the mystery. For reference, we also compare our results with those of voyage-context-3 in this setting.

Table~\ref{tab:detective_qa} highlights two key findings.
First, our SitEmb (QA) model achieves a substantial advantage in answer evidence recall over all other models, directly contributing to its higher final answer accuracy.
Second, our SitEmb (QA+SA) model shows a clear advantage in clue recall. This metric reflects an important aspect of semantic association capability, as clues are often only loosely or implicitly related to the question.

\begin{table*}[t]
  \centering
  \small
  \resizebox{\columnwidth}{!}{
  \begin{tabular}{l|ccc|ccc|ccc}
    \toprule
    \multirow{3}{*}{\textbf{Model}}& \multicolumn{3}{c}{\bf Answer Recall}& \multicolumn{3}{c}{\bf Clue Recall} & \multicolumn{3}{c}{\bf Final Accuracy}\\
    \cmidrule(lr){2-4}\cmidrule(lr){5-7}\cmidrule(lr){8-10}
     & Top-3 & Top-5 & Top-10& Top-3 & Top-5 & Top-10& Top-3 & Top-5 & Top-10\\
       \midrule
       voyage-context-3 & 36.1 & 46.8 & 63.3 & 24.8 & 33.8 & 48.1 & 68.7 & 73.5 & 79.8\\
       Qwen3 (out-of-box) & 29.6 & 37.8 & 55.5 & 23.8& 31.9&46.5 & 62.5 & 68.7&73.2\\
       \midrule
       Qwen3 (QA) & 35.8 & 50.5 & 66.4 & 23.7&33.0&48.0 & 70.5 & 78.2& 81.8\\
       SitEmb-v1.5-Qwen3 (QA)&\bf 42.5 & \bf 54.5 & \bf 69.3 & 24.6 & 34.0 & 49.2 & \bf 73.2 & \bf 78.7 & \bf 82.3\\
       SitEmb-v1.5-Qwen3 (QA+SA) & 29.4 & 41.3 & 56.7 & \bf 26.9 & \bf 36.4 & \bf 51.2 & 66.5 & 74.2 & 78.3\\
    \bottomrule
  \end{tabular}
  }
  \caption{{Study on the effects of improved retrieval on the DetectiveQA dataset, which provides evidence passage annotations.}}
  \label{tab:detective_qa}
\end{table*}

\section{Conclusion}
This paper introduces the \emph{situated embedding} models, which encodes a chunk’s surrounding contextual information directly into its embedding, enabling a deeper understanding of the chunk itself. Experiments across multiple long-context understanding tasks demonstrate that situated embeddings provide an effective alternative approach to contextual retrieval, and our proposed model serves as a strong first step in advancing this direction.

\section*{Limitations}
While our experiments on several use cases highlight the advantages of embedding models with enhanced semantic association capabilities, results on broader applications are mixed. At this stage, training with QA-only data achieves a better overall balance.
This suggests that semantic association exists along a spectrum from direct relevance to abstract and implicit relations. To excel across diverse scenarios, a model must be able to adaptively control its degree of divergence. Achieving this poses challenges for our current LoRA fine-tuning regime, which has limited capacity, and calls for new training objectives that explicitly encourage controllable association through instruction following.

Another limitation of our current work is that the models are primarily optimized for narrative data. In future work, we plan to construct training data from a broader range of domains to improve generalization.

\bibliography{custom}
\bibliographystyle{colm2025_conference}

\appendix
\section{Additional Training Details of Our Situated Embedding Model}
\label{app:training}
In this section, we describe additional details on how we attempt to train the first situated embedding model. 

\subsection{Model Initialization}
Before the residual learning process described in~\sref{sec:method}, we initialize two models, $\Theta^b$ and $\Theta^s$. While $\Theta^s$ is directly initialized from the BGE-M3 embedding model, we perform a prior training step on $\Theta^b$ to facilitate more effective residual learning.

Specifically, we initialize $\Theta^b$ from the same BGE-M3 embedding model as $\Theta^s$. For each query–chunk pair in the training data, we treat the chunk as the positive sample and randomly sample 10 negative chunks from other chapters of the same book. We then obtain the query embedding $\mathbf{q}^b$ and chunk embedding $\mathbf{c}^b$ from $\Theta^b$, and train $\Theta^b$ using the margin-based loss defined in Eq.~\ref{eq1}, applied solely to this model. This prior training stage familiarizes $\Theta^b$ with the task of retrieving book chunks based on user notes, thereby providing a more informative foundation for the subsequent residual learning phase.

\subsection{Training Configurations}
All training procedures in this paper follow a consistent configuration. For SitEmb-v1-M3, we use a learning rate of 2e-5 and a weight decay of 5e-2. The batch size is set to 80, and the maximum input sequence length is 8192 tokens, which corresponds to the input limit of BGE-M3. During training, we employ the development set introduced in~\sref{sec:method} for early stopping. The model is evaluated on this set every 180 training steps, and training is terminated once both the training loss and development performance converge. The margin and temperature values used in the loss function are both set to 0.1.
 All experiments referring to SitEmb-v1-M3 are conducted using two NVIDIA A100 80G GPUs.

For the training of SitEmb-v1.5-Qwen3, we equip Qwen3-Embedding-8B~\citep{zhang2025qwen3} with a Low-Rank Adaptation~\citep{lora}. The rank is set to 128, the alpha is set to 256, and adapters are attached to the query/key/value/output projections in multi-head attention modules, whose dropout rate is set to 0.05. The training schedule moves on using the cosine LR, warming up at the first 10$\%$ steps, whose learning rate is set to 1e-4. 

Unlike SitEmb-v1-M3, which first trains a chunk-only encoder and then residually trains a situated encoder with the chunk-only one frozen, we fully utilize the causal-masking feature of the decoder in Qwen3-Embedding (i.e., the unidirectional feature that history tokens cannot access future tokens), and train the context-only and situated settings at the same time. Specifically, for the chunk and context encoding, we concatenate them into one sequence in which the chunk comes first and is followed by the context. Due to the unidirectionality, the chunk can only see itself (i.e., the context-only setting), and the context can realize the situated chunk (i.e., the situated setting). The sequence is formed as ''[CHUNK]<|endoftext|>\textit{The context in which the chunk is situated is given below. Please encode the chunk by being aware of the context. Context:$\setminus$n}[CONTEXT]<|endoftext|>``. In this way, the chunk embedding and the situated embedding are obtained by the last pooling of extracting the embedding of the first and the second ''<|endoftext|>``. We denote the chunk embedding as $\mathbf{c}^b$, the situated embedding as $\mathbf{c}^s$, and the query embedding as $\mathbf{q}$. To co-train the two settings, the contrastive learning loss is computed using the scores sim($\mathbf{q},\mathbf{c}^b$), sim($\mathbf{q},\mathbf{c}^s$), sim($\mathbf{q},\mathbf{c}^b$) + sim($\mathbf{q},\mathbf{c}^s$), and the temperature value of 0.01. 

Furthermore, we follow~\cite{bge-m3,tevatron} to broadcast the computed scores of batches on every GPU to reach a bigger batch size. The batch size per GPU is set to 5, and we use 8 pieces of NVIDIA A800 80G GPUs. For every query, we sample a positive chunk plus 13 negative chunks from the same book. In this way, each query can see 8 * 5 * 14 = 560 chunks at a step. Additionally, the accumulation step is set to 4, and the model will be trained for 5 epochs. The best checkpoint is picked by the result on the NDP-v1 dev set per epoch, which is always from epoch 2. Therefore, we use the checkpoint saved at epoch 2 by default. All evaluating experiments referring to SitEmb-v1.5-Qwen3 are conducted using a piece of NVIDIA A100 40G GPU in the data type of bfloat16.

\section{Full Results Decomposed to Books}
\label{app:full_results}

\subsection{Books in the Evaluation Dataset}
\label{app:books}
Following the process described in~\sref{sec:dataset}, we select 7 books from the PlotRetrieval dataset to construct our evaluation set. The names of these books, along with the corresponding numbers of queries and candidate chunks, are summarized in~\tref{tab:statistics}. 

Note that for some English books, the PlotRetrieval dataset includes multiple Chinese translation versions, treating each version as a distinct book. We adopt the same setting and denote the three translation versions of \emph{Notre-Dame de Paris} in the 7 selected books as v1, v2, and v3, respectively. Among them, NDP-v1 is the version used in~\tref{tab:ndp_exp}.

\begin{table}[t]
  \centering
  \small
  \setlength{\tabcolsep}{1mm}{
  \begin{tabular}{lcc}
    \toprule
     \textbf{Book} & Queries & Candidates \\
    \midrule[0.5pt]
     \emph{Notre-Dame de Paris} (NDP)-v1&510&1288 \\ 
     \emph{Notre-Dame de Paris} (NDP)-v2&153&1369 \\ 
     \emph{Notre-Dame de Paris} (NDP)-v3&146&1347 \\ 
     \emph{Crime and Punishment} (C\&P)&134&1639 \\ 
     \emph{The Adventures of Tom Sawye} (TATS)&173&154 \\ 
     \emph{The Red and the Black} (TRB)&144&1294 \\ 
     \emph{Tess of the d’Urbervilles} (TDU)&134&1093 \\ 
    \bottomrule
  \end{tabular}
  \caption{Statistics of books in the evaluation dataset.}
  \label{tab:statistics}
  }
\end{table}

\begin{table*}[t]
  \centering
  \small
  \setlength{\tabcolsep}{3mm}{
  \begin{tabular}{llccc}
    \toprule
    \textbf{Model} & \textbf{Book} & @10 & @20 & @50 \\
    \midrule[0.5pt]
    \multirow{8}{*}{M3 (out-of-box)} 
      & NDP-v1 & 42.55 & 53.33 & 66.51 \\
      
      & NDP-v2 & 33.22 & 44.18 & 60.62 \\
      
      & NDP-v3 & 43.79 & 51.63 & 64.38 \\
      & C\&P & 21.64 & 24.63 & 45.15 \\
      & TATS & 38.73 & 47.40 & 63.01 \\
      & TRB & 23.61 & 34.72 & 47.22 \\
      & TDU & 26.87 & 34.33 & 44.03 \\
    \midrule[0.5pt]
    & Avg & 32.92 & 41.46 & 55.85 \\ 
    \midrule[0.5pt]
    \multirow{8}{*}{\makecell[l]{M3 (SA)}} 
      & NDP-v1 & 48.79 & 58.45 & 73.29 \\
      
      & NDP-v2 & 46.58 & 56.16 & 67.81 \\
      
      & NDP-v3 & 50.33 & 62.09 & 73.53 \\
      & C\&P & 36.57 & 42.91 & 54.85 \\
      & TATS & 42.20 & 56.94 & 70.81 \\
      & TRB & 36.80 & 43.06 & 58.33 \\
      & TDU & 38.81 & 50.75 & 63.43 \\
    \midrule[0.5pt]
    & Avg & 42.87 & 52.91 & 66.01 \\ 
    \midrule[0.5pt]
    \multirow{8}{*}{\makecell[l]{Res-M3 (SA)}} 
      & NDP-v1 & 49.30 & 58.73 & 71.93 \\
      
      & NDP-v2 & 48.29 & 54.79 & 66.78 \\
      
      & NDP-v3 & 50.33 & 59.15 & 72.22 \\
      & C\&P & 35.45 & 39.93 & 52.99 \\
      & TATS & 45.66 & 56.36 & 72.25 \\
      & TRB & 35.42 & 40.97 & 59.72 \\
      & TDU & 39.55 & 52.24 & 62.69 \\
    \midrule[0.5pt]
    & Avg & 43.43 & 51.74 & 65.51 \\ 
    \midrule[0.5pt]
    \multirow{8}{*}{\makecell[l]{SitEmb-v1-M3 (SA, No Res)}} 
      & NDP-v1 &48.76&58.53&72.22 \\
      
      & NDP-v2 &47.26&57.88&69.86 \\
      
      & NDP-v3 &51.31&63.07&74.84 \\
      & C\&P  &37.31&45.90&57.09 \\
      & TATS  &49.42&59.25&76.01 \\
      & TRB  &33.33&47.22&64.58 \\
      & TDU &39.55&52.99&67.91 \\
    \midrule[0.5pt]
    & Avg &43.85&54.98&68.93 \\ 
    \midrule[0.5pt]
    \multirow{8}{*}{\makecell[l]{SitEmb-v1-M3 (SA)}}
      & NDP-v1 & 50.85 & 60.57 & 76.40 \\
      
      & NDP-v2 & 48.97 & 57.88 & 70.89 \\
      
      & NDP-v3 & 51.63 & 64.71 & 74.84 \\
      & C\&P & 36.94 & 44.40 & 57.84 \\
      & TATS & 47.69 & 60.40 & 75.14 \\
      & TRB & 38.19 & 47.92 & 63.19 \\
      & TDU & 41.79 & 53.73 & 66.42 \\
    \midrule[0.5pt]
    & Avg & \textbf{45.15} & \textbf{55.66} & \textbf{69.25} \\ 
    \bottomrule
  \end{tabular}
  \caption{{Full results of SitEmb-v1-M3 on book plot retrieval.}}
  \label{tab:main_full_m3}
  }
\end{table*}

\begin{table*}[t]
  \centering
  \small
  \setlength{\tabcolsep}{3mm}{
  \begin{tabular}{llccc}
    \toprule
    \textbf{Model} & \textbf{Book} & @10 & @20 & @50 \\
    \midrule[0.5pt]
    \multirow{8}{*}{Qwen3-Embedding (out-of-box)} 
      & NDP-v1 & 51.20 &	61.63 	&76.10 \\
      
      & NDP-v2 & 38.36 &	47.95& 	69.86 \\
      
      & NDP-v3 & 50.65 &60.78& 	76.14  \\
      & C\&P & 35.82 &	43.66& 	60.45  \\
      & TATS & 49.71 &	60.69& 	75.14  \\
      & TRB & 43.06 &	50.00 &	65.97  \\
      & TDU & 36.19 &	45.90 &	62.69  \\
    \midrule[0.5pt]
    & Avg & 43.57 &	52.94 &	69.48  \\ 
    \midrule[0.5pt]
    \multirow{8}{*}{\makecell[l]{Qwen3 (QA)}} 
      & NDP-v1 & 61.66	&69.06&	79.54 \\
      
      & NDP-v2 & 52.74	&60.62	&78.42 \\
      
      & NDP-v3 &54.9	&70.59	&83.66 \\
      & C\&P &40.67	&52.61	&64.93 \\
      & TATS & 58.09	&65.61	&77.75 \\
      & TRB & 46.53	&56.25	&68.75 \\
      & TDU & 42.54	&50.74	&64.18 \\
    \midrule[0.5pt]
    & Avg & 51.02 &	60.78 &	73.89 \\ 
    \midrule[0.5pt]
    \multirow{8}{*}{\makecell[l]{Qwen3 (QA+SA)}} 
      & NDP-v1 & 66.81 &	74.36 &	84.32 \\
      
      & NDP-v2 & 67.47 &	75.68 &	84.59 \\
      
      & NDP-v3 & 65.03 &	71.90 &	87.91 \\
      & C\&P & 49.63 &	61.57 &	73.51  \\
      & TATS & 60.69 &	73.12 &	84.97  \\
      & TRB & 56.94 &	63.54 &	73.96  \\
      & TDU & 55.97 &	61.94 &	73.88  \\
    \midrule[0.5pt]
    & Avg & 60.36 &	68.87 &	80.45  \\ 
    \midrule[0.5pt]
    \multirow{8}{*}{\makecell[l]{SitEmb-v1.5-Qwen (QA)}} 
      & NDP-v1 &63.32&	72.75&	83.59 \\
      & NDP-v2 &58.90&	68.84&	80.82 \\
      & NDP-v3 &59.48&	70.59&	84.64 \\
      & C\&P  &44.40&	54.10&	76.49 \\
      & TATS  &58.96&	70.23&	83.24 \\
      & TRB  &45.14&	57.64&	71.53\\
      & TDU &44.78&	50.75&	70.15 \\
    \midrule[0.5pt]
    & Avg &53.57&	63.56&	78.64 \\ 
    \midrule[0.5pt]
    \multirow{8}{*}{\makecell[l]{SitEmb-v1.5-Qwen (QA+SA)}}
      & NDP-v1 & 68.98& 	79.32& 	86.68 \\
      & NDP-v2 & 71.23& 	79.45&	89.73 \\
      & NDP-v3 & 66.34& 	79.41& 	87.58 \\
      & C\&P & 58.58& 	68.28& 	76.49  \\
      & TATS & 65.61& 	73.70& 	84.10  \\
      & TRB & 50.00& 	62.50& 	77.43  \\
      & TDU & 60.45& 	67.16& 	76.87 \\
    \midrule[0.5pt]
    & Avg & \bf 63.03& 	\bf 72.83& 	\bf 82.70  \\ 
    \bottomrule
  \end{tabular}
  \caption{{Full results of SitEmb-v1.5-Qwen on book plot retrieval.}}
  \label{tab:main_full_qwen}
  }
\end{table*}

\section{Details on Running Embedding Models}
\label{app:run_model}

For all non-LLM embedding models (i.e., BGE-M3 and Jina-v3), we directly use the models to encode queries, chunks, and situated context, with the maximum input length set to 8192 tokens.

For E5-Mistral, GTE-Qwen2, and voyage-context-3, we follow the official encoding guidelines provided at~\url{https://huggingface.co/intfloat/e5-mistral-7b-instruct}, ~\url{https://huggingface.co/Alibaba-NLP/gte-Qwen2-7B-instruct}, and~\url{https://docs.voyageai.com/docs/contextualized-chunk-embeddings#quickstart}, respectively. In both cases, we prepend a one-sentence instruction to each query as required, as illustrated in~\fref{fig:e5_prompt}.

For NV-Embed-v2, we adopt the same input format as E5-Mistral and GTE-Qwen2 when encoding queries. For chunk encoding in the chunk-only setting of~\tref{tab:ndp_exp}, we omit instructions, consistent with the E5-Mistral and GTE-Qwen2 setups. In all other settings in~\tref{tab:ndp_exp} where additional context is included, we follow the official prompt format of NV-Embed-v2~(\url{https://huggingface.co/nvidia/NV-Embed-v2}), as shown in~\fref{fig:nv_prompt}.

When running the latter two columns of experiments in~\tref{tab:ndp_exp}, we append the situated context or situated summary to each chunk using the delimiters ``</s>'' and ``\textbackslash n\textbackslash n'', respectively. The concatenated sequence is then treated as a new chunk and encoded as a whole. 

\begin{figure*}
  \begin{tcolorbox}
  \textbf{Instruct}:\\
  Given a user note query, retrieve the passages that are most relevant to the content or context described in the query.\\\\
  \textbf{Query}:\\
  \{QUERY\}

    \end{tcolorbox}
    \caption{The query format of E5-Mistral and GTE-Qwen2}
    \label{fig:e5_prompt}
\end{figure*}

\begin{figure*}
  \begin{tcolorbox}
 Your task is to embed passages for retrieval. Your input consists of the target passage and its context. You need to find relevant information from the context to enhance the target passage embedding such that it captures the meanings of the passages situated within the context.
\\\\
  \textbf{context}: \\
  \{CONTEXT\}\\\\
  \textbf{passage}: \\
  \{PASSAGE\}

    \end{tcolorbox}
    \caption{Prompt for NV-Embed-v2.}
    \label{fig:nv_prompt}
\end{figure*}
\end{document}